\documentclass[letterpaper, 10 pt, conference]{ieeeconf}  

\IEEEoverridecommandlockouts                              

\usepackage{graphicx}
\usepackage{amsmath}
\usepackage{amssymb}
\usepackage{booktabs}
\usepackage{amsmath}
\usepackage{paracol}
\usepackage{empheq}
\usepackage[table]{xcolor}
\usepackage{tabularx}
\usepackage{multirow}
\usepackage[pagebackref,breaklinks,colorlinks]{hyperref}
\usepackage{paracol}
\usepackage[capitalize]{cleveref}
\crefname{section}{Sec.}{Secs.}
\Crefname{section}{Section}{Sections}
\Crefname{table}{Table}{Tables}
\crefname{table}{Tab.}{Tabs.}
\usepackage{tikz,xcolor}
\usepackage{gensymb}

\definecolor{lime}{HTML}{A6CE39}
\DeclareRobustCommand{\orcidicon}{%
    \begin{tikzpicture}
    \draw[lime, fill=lime] (0,0) 
    circle [radius=0.16] 
    node[white] {{\fontfamily{qag}\selectfont \tiny ID}};
    \draw[white, fill=white] (-0.0625,0.095) 
    circle [radius=0.007];
    \end{tikzpicture}
    \hspace{-2mm}
}

\newcommand{\ibrahim}{\href{https://orcid.org/0000-0002-5376-2477}{\orcidicon}}
\newcommand{\naveed}{\href{https://orcid.org/0000-0003-3406-673X}{\orcidicon}}
\newcommand{\ajmal}{\href{https://orcid.org/0000-0002-5206-3842}{\orcidicon}}

\title{\LARGE \bf
Slice Transformer and Self-supervised Learning for 6DoF Localization 
in 3D Point Cloud Maps
}

\author{Muhammad Ibrahim\ibrahim$^{1}$,  Naveed Akhtar\naveed$^{1}$,  Saeed Anwar$^{2}$, Michael Wise$^{1}$ and  Ajmal Mian\ajmal$^{1}$ 
\thanks{Professor  Ajmal  Mian  is  the  recipient  of  an  Australian Research Council Future Fellowship Award (project number FT210100268) funded by the Australian Government.
Dr.~Naveed Akhtar is recipient of an Office of National Intelligence National Intelligence Postdoctoral Grant (project number NIPG-2021-001)
funded by the Australian Government.}
\thanks{$^{1}$Department of Computer Science, The University of Western Australia.
        {\tt\small muhammad.ibrahim@research, naveed.akhtar@, michael.wise@, ajmal.mian@) uwa.edu.au}
        }%
\thanks{$^{2}$King Fahad University of Petroleum and Minerals (KFUPM), Dhahran, KSA, 
        {\tt\small saeed.anwar@kfupm.edu.sa}}%
        }

\begin{document}

  \makeatletter
\let\@oldmaketitle\@maketitle
\renewcommand{\@maketitle}{\@oldmaketitle  
\label{map2}
\centering
 \includegraphics[width=0.9\columnwidth*2]{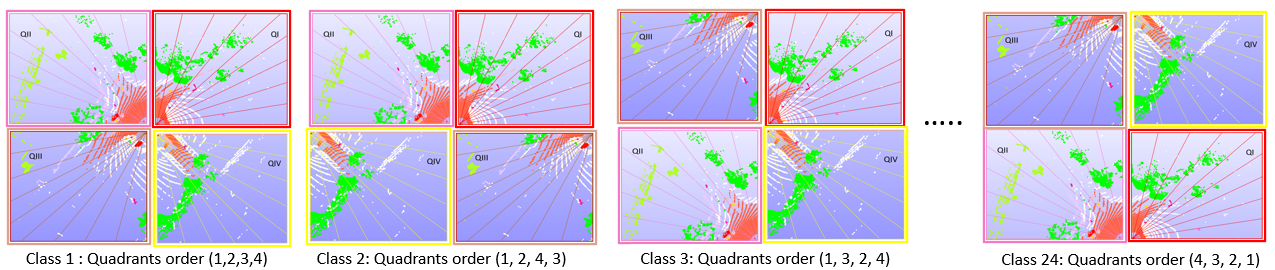}\\
 \vspace{-2mm}
    {\footnotesize{Fig.~1: The proposed self-supervised pretext task shuffles 3D point cloud quadrants to create pseudo-labels. Quadrant permutations result in 24 labels.  36 slices are generated from a single raw frame, covering 30$^{\circ}$ rotation that includes a 20$^{\circ}$ overlap with the neighboring slices - overlap not shown.
    \vspace{-3mm}
    }}}

\maketitle 

\thispagestyle{empty}
\pagestyle{empty}

\begin{abstract}
Precise localization is critical for autonomous vehicles. We present a self-supervised learning method that employs transformers for the first time for the task of outdoor localization using LiDAR data. We propose a pre-text task that reorganizes the slices of a $360^\circ$ LiDAR scan to leverage its axial properties. Our model, called Slice Transformer, employs multi-head attention while systematically processing the slices. To the best of our knowledge, this is the first instance of leveraging multi-head attention for outdoor point clouds. We additionally introduce the Perth-WA dataset, which provides a large-scale LiDAR map of Perth city in Western Australia, covering $\sim$4km$^2$ area. Localization annotations are provided for Perth-WA.  The proposed localization method is thoroughly evaluated on Perth-WA and Appollo-SouthBay datasets. We also establish the efficacy of our self-supervised learning approach for the common downstream task of object classification using ModelNet40 and ScanNN datasets. The code and Perth-WA data will be publicly released.


\end{abstract}

\section{INTRODUCTION}
Six degrees of freedom (6DoF) localization of vehicles is a key task for  autonomous driving. Satellite-based localization lacks the required precision and does not work inside cities with tall structures, bridges and tunnels. To achieve the required level of precision in the mentioned scenarios, 3D LiDAR-based localization is an optimal choice.

A general approach towards LiDAR-based localization is to construct an offline 3D map and query the map with LiDAR frames during online navigation. Conventional techniques under this paradigm~\cite{segal2009generalized, stoyanov2012fast} leverage frame registration. However, due to their large  computational requirements, they are unable to provide a  practical solution. More recently, deep learning based techniques have shown promising results. Among these methods, matching deep learning features of an input frame with the deep features of a pre-computed map is a viable solution~\cite{yew20183dfeat, li2021robust}. Nevertheless, such methods also require post-processing with registration to achieve the desired precision level~\cite{Elbaz_2017_CVPR}. 

To contain the prediction latency within practical limits, the state-of-the-art methods  compress the map into a neural model, and directly localize vehicles by regressing its 6DoF pose over the map~\cite{wang2021pointloc}. For 3D point clouds, this demands high-fidelity representation learning by the model because of the unstructured nature of the data. Though effective, conventional deep learning methods based on multi-layer perceptrons~\cite{qi2017pointnet} and convolutional networks \cite{komarichev2019cnn} still fall short on the accuracy required for the critical task of 6DoF localization with LiDAR frames.  

Transformers~\cite{vaswani2017attention} have recently surpassed conventional deep learning methods in performance. However, their extension to point cloud data is not straight-forward. In fact, currently, there is no widely known technique that can leverage the key strength of transformers, i.e., multi-head attention, for outdoor point cloud data. An additional challenge is that due to their low inductive bias, transformers also require an even larger amount of data than convolutional networks.   

In this work, we make three major contributions to achieve highly precise 6DoF localization in 3D point clouds. First, we propose a pre-text task for self-supervised learning on point clouds, see Fig.~1. Our task exploits the intrinsic axial nature of the LiDAR data by systematic slicing of each frame. The slices are shuffled to allow pre-training on a very large mount of data in a self-supervised way, which is especially conducive for learning effective Transformer models. Second, we propose a first-of-its-kind based Transformer model that enables leveraging multi-head attention for outdoor point clouds. Specifically suited to our pre-text task, the model processes a LiDAR frame by slicing it, hence termed Slice Transformer. Third, we introduce a 3D LiDAR map of the Perth city in Western Australia that covers $\sim4$ km$^2$ of the Central Business District (CBD), providing annotations for 6DoF localization as well as additional frames for self-supervised learning.

We establish the baseline results for 6DoF localization on our dataset with the proposed method and PointLoc~\cite{wang2021pointloc}. We also provide benchmarking of our technique on an existing Appolo-Southbay dataset~\cite{L3NET_2019_CVPR}. Moreover, leveraging the backbone of our model, we demonstrate the effectiveness of our model for point cloud object classification task with ModelNet40~\cite{wu20153d} and ScanObjectNN~\cite{Uy_2019_ICCV} datasets.

\begin{figure*}
\begin{center}
\includegraphics[width=1\columnwidth*2]{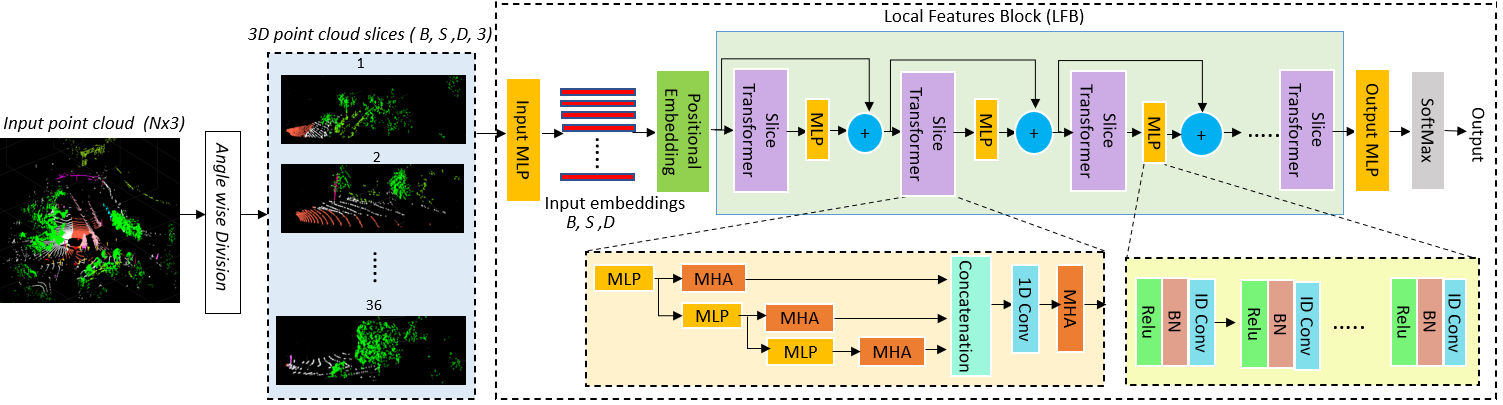}
\end{center}
\vspace{-5mm}

\caption{Schematics of the proposed method. A 3D point cloud frame of $N$ points is divided into 36 slices with overlap. Embedding for each slice is created for tokenization and positional encoding is added. Local Feature Block (LFB) is proposed to enable multi-headed attention (MHA) over multi-layer perceptron (MLP) embeddings of input features. This process forms the core of Slice Transformer. Processing the features with multiple Slice Transformer blocks, MLPs and skip connections, the output is processed with a softmax layer for the pre-text classification task. The + sign stands for addition.}
\label{fig:classifier}
\vspace{-6mm}
\end{figure*}

\section{Related Work}
\vspace{-1mm}
6DoF localization is an important task for self-driving vehicles~\cite{elhousni2020survey, gao2021fully}. Whereas conventional methods for matching point cloud frames for such a task have used registration techniques~\cite{segal2009generalized, kovalenko2019sensor}, more recent works focus on exploiting deep learning  to relate an input frame to a 3D map~\cite{wang2021pointloc, L3NET_2019_CVPR, nubert2021self, dube2018segmap}. Among these methods, there are contributions that compress the map into a neural model and use that model as a 6DoF pose predictor for the vehicle~\cite{wang2021pointloc, L3NET_2019_CVPR}. Using raw LiDAR frames, this prediction is particularly challenging due to the unstructured nature of the data, which conflicts with the high precision requirements of the task. Hence, it is imperative to continuously improve the neural models for this task along with the progress of deep learning.

Recent literature has seen Vision Transformer~\cite{vaswani2017attention} and its variants, e.g., \cite{liu2021swin}, \cite{chen2021crossvit}, \cite{yuan2021tokens}, \cite{touvron2021training}, \cite{strudel2021segmenter}, \cite{chen2021pre}, to outperform their convolutional and multi-layer perceptron based counterparts on a variety of  tasks. Consequently, Transformer architectures have also started to emerge for point cloud processing. For instance, \cite{guan2021m3detr} and \cite{misra2021end} employ transformers for 3D object detection, whereas \cite{zhao2021point} developed a method for point cloud segmentation. However, due to the complex nature of outdoor LiDAR data, Transformer models are yet to establish themselves on  outdoor benchmarks.

In general, Transformer architectures require a huge amount of training data to learn effective models. In the domain of  point clouds, this need is partially filled with self-supervised learning~\cite{yu2022point, fu2022pos, wang2022p2p}. However, the  pre-text tasks are designed keeping in mind the downstream tasks of object classification and segmentation. Representative examples of pre-text tasks in point cloud domain include self-shape correction of 3D CAD models~\cite{chen2021shape}, self-domain adaptation~\cite{Achituve_2021_WACV} and  point cloud rotation~\cite{9320440}. In general, the issue with the self-supervised methods for point cloud processing  is that the pre-training is  performed  either on images~\cite{wang2022p2p}, 3D synthetic data~\cite{fu2022pos, Achituve_2021_WACV} or indoor point clouds~\cite{yu2022point}, which is not particularly helpful for outdoor downstream applications like localization.



Contemporary self and un-supervised  deep learning based point cloud techniques employ CNNs and MLPs. Nubert et al.~\cite{nubert2021self} proposed a self-supervised  method to predict poses from input LiDAR frames. Similarly, an un-supervised projection  method was proposed by Cho et al.~\cite{cho2020unsupervised} to predict the poses from input. Both methods exploit structured 2D CNNs to process 3D point clouds indirectly. SelfVoxeLO~\cite{xu2020selfvoxelo}  exploits 3D CNNs to process the raw point cloud directly for pose estimation. Similarly, UnPWC-SVDLO~\cite{tu2022unpwc} utilizes scene flow estimation network PointPWC~\cite{wu2019pointpwc} as a backbone  for its un-supervised LiDAR odometry.

Leveraging the representation prowess of transformers for point cloud analysis is still under-explored. 
Along this nascent direction, encouraging results have started to emerge~\cite{zhou2021ndt, shan2021ptt}. However, the complexity of data, especially in the outdoor domain, has still not allowed researchers to exploit the key strength of transformers, i.e., multi-head attention, in their techniques. Current methods are limited to simple indoor scenes and synthetic objects. Our work fills this gap by enabling multi-head attention for outdoor tasks on challenging LiDAR scans. Moreover, we introduce a pre-text learning task especially suited to outdoor LiDAR data and present the first instance of outdoor localization with transformers in the domain of point clouds.

\section{SELF-SUPERVISED LEARNING}
\label{sec:SSL}
\vspace{-2mm}
\subsection{Pre-text Task }
\label{sec:Pretext}
 \vspace{-2mm}
Due to the complex nature of outdoor LiDAR data, it is imperative to induce models with a large amount of training samples. However, annotating a large number of LiDAR frames for any downstream task can become prohibitive. To side-step the issue, we propose a pre-text task to automate the labeling process with pseudo-labels. The central idea of the proposed pre-text task is illustrated in Fig.~1. 

We first divide a LiDAR frame axially into 36 slices. These slices are split into four quadrants of 9 slices each. We generate a new class label by shuffling the quadrants, resulting in 24 pseudo-labels. Each slice contains 10$^{\circ}$ of new region and a 20$^{\circ}$ overlap with its neighboring slices. Our treatment of a frame in slices is governed by the tokenization of input required by the Transformer architecture. The slices naturally result in well-defined tokens. We provide the architectural details of our network in Section~\ref{sec:App}. Using the pre-text task, we used 250K raw LiDAR frames to pre-train our model using the standard cross-entropy loss. These frames are taken from the proposed Perth-WA dataset (see Sec.~\ref{sec:PerthWA}). The model is trained for 100 epochs on 240K raw frames and 10K frames are used for validation. Our model achieved around 95\% classification accuracy on the  validation set.

  \vspace{-2mm}
\subsection{Proposed Slice Transformer Network}
\label{sec:App}
 \vspace{-2mm}
We propose a Transformer architecture that leverages multi-head attention to process outdoor LiDAR frames. Illustrated in Fig.~\ref{fig:classifier}, our network consumes slices of a frame, which is inline with the pre-text task discussed in \ref{sec:Pretext}. We describe the major processes of the network below.
\vspace{0.5mm}
\subsubsection{Slice~Extraction}
A point cloud frame of size $\mathbb R^{N \times 3}$ is transformed to $\mathbb R^{S \times D \times 3}$ where $N$, $S$ and $D$ are the number of input points, number of slices and dimension of each slice respectively. For slice extraction, we first transform X, Y, Z values of in Cartesian coordinated to  Azimuth, Elevation, and Radius. Along the Azimuth, points falling in  $30^{\circ}$ slices are extracted. Then, a $10^{\circ}$  rotation is applied to the point cloud to extract a new slice. 
This allows slice extraction with a $20^{\circ}$ overlap. In our implementation, we filter out the less dense regions beyond 70m radius in a slice. 

\vspace{0.5mm}
\subsubsection{Input Embeddings Generation}
We transform 3D slices of size  $\mathbb R^{S \times D \times 3}$ into input emdeddings in  $\mathbb R^{S \times D}$. 
To that end, we employ three 2D CNN layers with the input channel sizes 36, 64 and 128. These layers are followed by four 1D CNN layers, with output channel sizes 128, 128, 64, and 36. These layers are packed into a block, which is called an Input MLP block in Fig.~\ref{fig:classifier} due to the dominant use of 1D CNN layers for projection purpose.

\vspace{0.5mm}
\subsubsection{Positional Embedding}
The positional information to the input embeddings in transformers  is known to  boost the network performance. Unlike the conventional `cosine' and `sine' positional encoding, we add learnable slice-wise   positional encodings for improved performance. This learned positional encoding is added to the input data. 
\vspace{0.5mm}
\subsubsection{Local Features Block (LFB)}
Local features  play a key role in  neural representation. 
We exploit both multi-headed attention (MHA) and CNNs to extract powerful features from the input point cloud. Local Feature Block (LFB) is the central component of our network that  extracts local features of a raw point cloud frame. It incorporates pairs of the  proposed Slice Transformer and an MLP in series as shown in Fig~\ref{fig:classifier}. Between the pairs, skip-connections are used. We  empirically selected three pairs of  slice transformers and  MLP in our network. 

\vspace{0.5mm}
\noindent{\bf{\textit{Slice Transformer}:}} This module processes a point cloud as a sequence of slices, hence called Slice Transformer. As shown in Fig.~\ref{fig:classifier}, it comprises four Multi-head Attentions  (MHAs) and three MLPs. The input and output of this block are in  $\mathbb R^{S \times D }$. An MHA is based on a Scaled Dot-Product Attention defined as 
 \vspace{-5mm}
\begin{flalign}\label{eq1}
\vspace{-3mm}
 \text{Atten}(q,k,v) &=  v~ \text{softmax}(\frac{q\cdot k^\intercal}{\sqrt{dk}}), 
\end{flalign}
where  Atten is the attention function of a single head; $q$, $k$, $v$ are query, key and value; $dk$ is key's dimension.  

In the case of MHA, keys, queries and values are projected multiple times to different learned linear transformations. The attention function is computed in parallel over all projections separately. The outputs  of these computations are concatenated. For efficiency, we use 16 heads with 128 dimensions each. Thus, the input to a single head attention function is $\mathbb R^{S \times 128 }$. 
MHA permits the network to attend useful features from different representation subspaces at various positions. 
In Slice Transformer, we place MLPs before MHA modules at various locations to extract useful and diverse local features of the input data. Each MLP consists of six 1D CNNs with Batch Normalization layers and ReLU activation functions. The input and output channel size for the CNNs are      $(36,64),(64,64),(64,128),(128,128),(128,128),(128,36)$. Mathematically, a single head performs the computation as shown in Eq.~(\ref{eq1}) while MHA performs concatenation operation over all heads computations, which can be expressed as 
\vspace{-2 mm}
\begin{equation}
    \\ \label{eq2}
 \text{MHA}(q,k,v) = \text{concat}(h1,h2,h3....hi)W^O,
\end{equation}

where $h_i$ denote attention heads and $W^O \in \mathbb R^{dk \times 2048}$ is the projection parameter matrix.

\begin{figure*}
\begin{center}
\includegraphics[width=1\columnwidth*2]{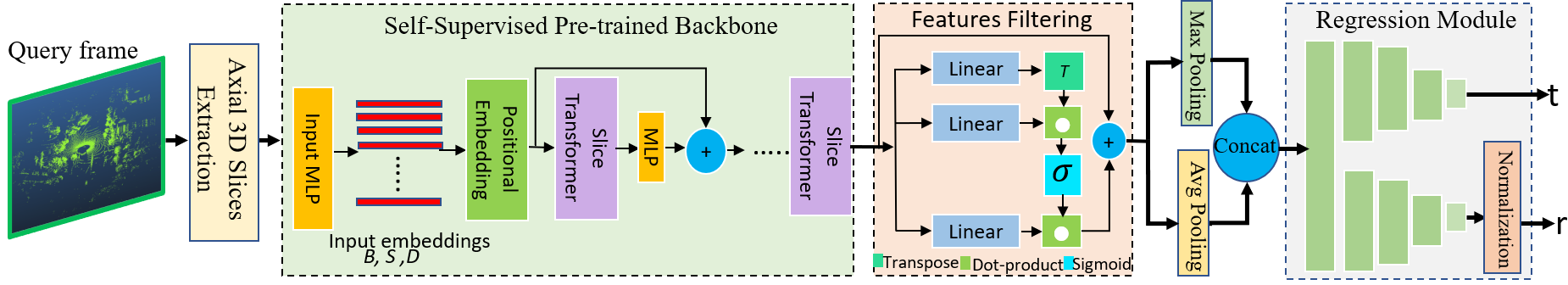}
\end{center}
\vspace{-6mm}

\caption{Proposed end-to-end 6DoF localization method that leverages the self-supervised pre-trained backbone followed by a regression stage that employs two-headed fully-connected sub-network to predict translation and rotation parameters for the query point cloud frame.
}
\label{fig:loc}
\vspace{-5mm}
\end{figure*}
\subsubsection{Output Classification}
Within the context of pre-text learning, this process is responsible for the final  prediction. It is implemented with  5 1D CNNs with kernel size $(1,1)$, a maxpooling layer, and two linear layers. The 1D CNNs map the input features  $\mathbb R^{S \times D }$ to  $\mathbb R^{1024 \times 2048 }$ and then maxpooling is employed  channel-wise. Finally, linear layers and softmax is applied to generate the final output. Using the conventional cross-entropy loss, the network is trained to predict pseudo-labels for the pre-text task or the true labels for classification- see Sec.~\ref{sec:CT}.  
\vspace{-3mm}
\section{LOCALIZATION}
\vspace{-1mm}
Self localization is a critical autonomous navigation task. We present a method for localization in 3D point cloud map with LiDAR input by building on the self-supervised model presented in the preceding section. Our technique is illustrated in Fig.~\ref{fig:loc}. It divides a query frame into slices following the method of  Sec.~\ref{sec:Pretext}, and extracts features for slices using a pre-trained backbone obtained using the technique in Sec.~\ref{sec:SSL}. The output of this module is further processed using the components described below.


\subsection{Features Filtering}
\vspace{-1mm}
Features extracted by the backbone network may also contain unwanted and noisy features. Inspired by \cite{wang2020atloc}, \cite{huang2019prior}, we devise a neural modeling based filter to cleanse  the unwanted features.
To that end,  we first estimate a mask of size $\mathbb R^{1 \times D }$ for the input slice features in $\mathbb R^{S \times D }$ with an MLP employing sigmoid activation. The mask is then broadcasted to the features  by applying dot product between the  mask and input features followed by normalization. This enables extraction of similar features from the input features map. 

\subsection{Max-pooling and Average Pooling}
\vspace{-1mm}
The main responsibility of these layers is to identify slice-wise maximum and average features. The max-pooling and average pooling generate outputs with size $\mathbb R^{D \times 1 }$, expecting  inputs of size $\mathbb R^{D \times S }$. Outputs from these layers are concatenated for further processing.

\subsection{Regression Module}
\vspace{-1mm}
The pre-trained model from Sec.~\ref{sec:SSL} is a classifier whereas 6DoF prediction is better formulated as a regression problem.  Therefore, we replace classification related modules with a Regression Module, while keeping the pre-trained backbone. The employed Regression Module consists of a common fully-connected (FC) layer at the initial stage, followed by two branches of four FC layers - see Fig.~\ref{fig:loc}. 
The output channel sizes in each branch of the FC layers are $1024,512,256,3$. The layers are initialized with  Xavier\_uniform distribution, and use ReLU activations. We also normalize the rotations branch of the sub-network to account for the relatively smaller  variations.

\subsection{Training Loss Function}
\vspace{-1mm}
Unlike other localization methods which are predominantly restricted to a single  loss  for both rotation and translation, we use $L_1$-loss for translation vector and Cosine Similarity Embedding loss for the rotation vector. We combine  these  losses with learnable balancing factors $\alpha$  and $\beta$, as the net batch loss  in Eq.~(\ref{eq3}).
\begin{flalign}\label{eq3}
 \mathcal L &=   \| t - t'\| e^{\alpha} +  \alpha +  ( 1 - \cos(r - r'))e^{\beta} +\beta,
\end{flalign}
where  $(t',r')$, $(t,r)$ are the ground-truth and predicted poses i.e., translations $t$ (along X, Y, Z directions) and rotations $r$ (in yaw, roll, pitch).

\begin{figure}
\includegraphics[width=1\columnwidth]{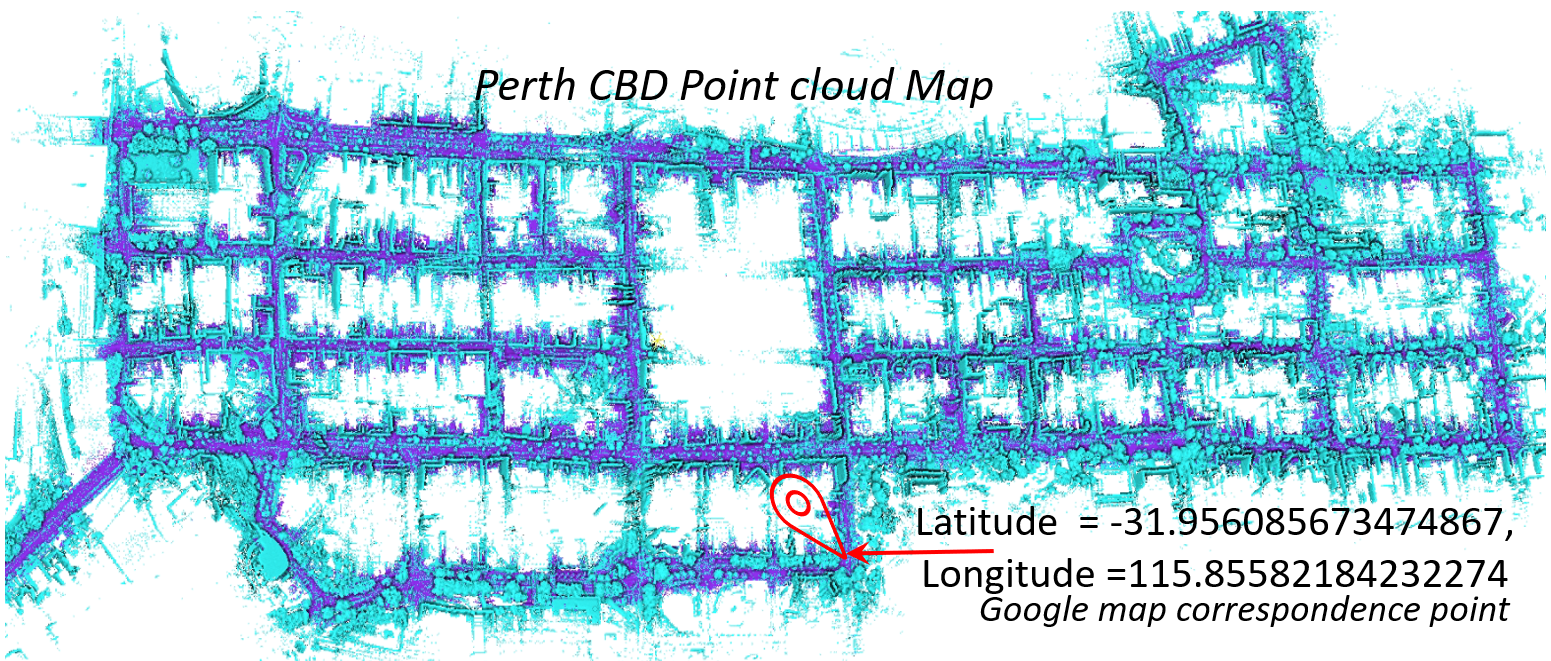}
\vspace{-4mm}

\caption{Constructed 3D point cloud map of Perth CBD, WA. The map shown binary annotations of roads and other objects. 
}
\label{fig:map}
\vspace{-7mm}
\end{figure}

\section{Perth-WA Dataset}
\label{sec:PerthWA}
\vspace{-2mm}
Another major contribution of this work is Perth-WA dataset that provides 6DoF annotations for localization. The data comprises a LiDAR map of 4km$^2$ region of Perth Central Business District (CBD) in Western Australia as shown in Fig.~\ref{fig:map}. The scenes contain  commercial structures, residential areas, food streets, complex routes, and hospital building etc. The data was collected in three different two-hour sessions under day/night conditions with sunny and cloudy weather.  Unlike the existing related dataset, Apollo-SouthBay~\cite{L3NET_2019_CVPR} and  Oxford Radar RobotCar Dataset~\cite{RadarRobotCarDatasetICRA2020}, Perth-WA dataset annotations do not rely on Inertial Measurement Unit (IMU). Instead, the labeling comes directly from the LiDAR frames themselves, as explained shortly.

In constructing a large 3D point cloud map, accumulation of registration errors is a major problem. 
This causes a drift in the map. To overcome that, we collected the data in closed 
loops. It is known that registration of closed loop LiDAR frames reduces error accumulation  drastically~\cite{chng2020monocular}. Our  map  is constructed in an offline process,  which involves extraction of loops from the scanned data, filtering out non-static objects, registration and merging the consecutive frames in a loop, followed by merging all the sub-maps into a single map. We further post-process the map 
to remove redundant points. In map construction, we leverage 3D normal-distribution transform (3D-NDT) algorithm~\cite{magnusson2009three} for registering the  LiDAR frames. 
There are total nineteen sub-maps created from raw LiDAR frames which are merged to create a single 3D point cloud map. We allow two types of regions with dense and sparse point clouds in the map, 
the latter to create a more challenging scenario for localization. 
For the sparse region of the map, we skip three consecutive frames in the collected data. 

To extract the ground-truth poses for Perth-WA dataset, we exploit the map creation process itself. Within a loop, a moving LiDAR frame is registered with a static point cloud, which generates a transformation matrix for the moving frame. To compute the transformation matrix of a frame, we multiply the previous frame's transformation to the transformation matrix of that frame. For instance, when the frames in a loop are $f_{1}, f_{2}, f_{3}, \dots f_{n}$, and their corresponding  transformation matrices are $ T_{1}, T_{2},T_{3}, \dots T_{n}$, then the ground truth transformation matrices for each frame can be computed as $(T_{0} \cdot T_{1}) = T_{g1}, (T_{g1} \cdot T_{2}) = T_{g2}, \dots (T_{gn-1} \cdot T_{n}) = T_{gn}$. The $T_{0}$ is a $4 \times 4$ identity matrix for the first loop, however, it is the  ground-truth transformation matrix of the last frame in a previous loop for any other loop. To compute the ground-truth transformation matrices for the next loop, we initialise the registration process from $T_{0}$ to compute the ground-truth of the first frame in that loop. After merging two loops, we refined the values of $T_{0}$ with  rotation and translation values to ensure that the frames align perfectly with the map. We use this strategy for all the loops to generate the ground-truth annotations with high precision. Finally, we convert the ground-truth transformation metrics to rotation angles (degrees) and translation vectors (meters) for all the  frames to provide annotations for training and testing the model. The Perth-WA dataset contains 30K frames with  6DoFs poses. 
 \vspace{1mm}
\section{Experiments}
\vspace{-2mm}
We evaluate the proposed method for localization and object classification tasks on benchmark datasets and compare with the state-of-the-art methods. Moreover, we also establish baseline results on the proposed Perth-WA dataset. For comparison, we choose existing localization and classification methods  based on their popularity and availability of the author-provided code for a fair evaluation.

\subsection{Setup} 
 \vspace{-2mm}
All experiments are performed on Ubuntu 18.04 operating system with PyTorch version 1.9 and using a single GeForce RTX 3090 GPU with 24GB memory. 
For the proposed method, each point cloud frame is divided into 36 slices, 30$^\circ$ each with 2048 points per slice and 20$^\circ$ overlap with the neighboring slices.

\begin{table}[t!]

\caption{Classification results (Acc \%) on ModelNet40 and ScanObjectNN. PC = Point Cloud, N = Normal.}
\vspace{-2mm}

\resizebox{1\columnwidth}{!}{
\begin{tabular}{l c  c c l}
 \hline
\rotatebox[origin=c]{0}{Method} & \rotatebox[ origin=c]{0} {Input}&      \rotatebox[origin=c]{0}{ ModelNet40} &  \rotatebox[origin=c]{0}{ScanNN}  &\\ \hline
PointNet++\cite{qi2017pointnet++} & PC &  90.7  & 77.9 & \\
DGCNN~\cite{mohammadi2021pointview} & PC &    89.0 & 78.4 & \\ 
SpiderCNN (4-layer)~\cite{xu2018spidercnn} & PC+N &  \bf{92.4} & 73.7 & \\

PointNet\cite{qi2017pointnet} & PC &  89.2  & 68.2 & \\
PointCNN\cite{li2018pointcnn} & PC+N &  92.2  & 78.5 & \\
3DmFV~\cite{8394990}& PC  & 91.4  & 63.0 & \\
Transformer~\cite{yu2022point} & PC &  91.4 & 77.2\\ 
Transformer+OcCo~\cite{wang2021unsupervised} & PC  & 92.1  & 80.4 & \\
NPCT ~\cite{guo2021pct} & PC  & 91.0  & - & \\
POS-BERT ~\cite{fu2022pos} & PC  &  92.1 & 83.2\\
Ours & PC  & \bf{92.4}  & \bf{84.5} & \\
\hline
\end{tabular}}
\label{tab:table3}
\vspace{-5mm}

\end{table}

\subsection{Classification  Task }
\label{sec:CT}
 \vspace{-1mm}
 For  classification, we directly fine-tune our self-supervised pre-trained model for the task. We present results on the popular ModelNet40~\cite{wu20153d} and ScanObjectNNN~\cite{Uy_2019_ICCV} datasets. 
 
\subsubsection {Datasets} 
ModelNet40~\cite{wu20153d} contains 3D CAD models of 40 object classes. We use the standard training  and test set in our experiments, comprising 9,843  and 2,468 samples, respectively. The ScanObjectNN dataset~\cite{Uy_2019_ICCV} comprises nearly 15K real-world object scans with occlusions and background.  It is a challenging point cloud dataset for classification task. It has 15 categories with 2,902 distinct object instances.  Each object in the dataset is defined by a list of 3D points ($x,y,z$ values), normal vector and color at each point, and class label. We perform experiments on the perturbed (PB-T50-RS) variant of the dataset.
 
\subsubsection {Implementation Details} 
In classification, we set the same training parameters for both datasets. We employ batch sizes of 16 and 4 for training and testing, respectively; and use Adam optimizer with learning rate 0.001 and weight decay 0.00001. We use the standard Cross Entropy loss and train the model for 60 epochs each on both datasets.
 
\subsubsection{Results on ModelNet40 and ScanObjectNN Datasets}
We compare our method with the state-of-the-art. Among the MLP and CNN based classification methods, we compare with the PointNet\cite{qi2017pointnet}, PointNet++\cite{qi2017pointnet++}, DGCNN~\cite{mohammadi2021pointview}, SpiderCNN~\cite{xu2018spidercnn}, PointCNN\cite{li2018pointcnn}, and 3DmFV~\cite{8394990}. We also compare with transformer based methods, which include Transformer~\cite{yu2022point}, Transformer+OcCo~\cite{wang2021unsupervised}, POS-BERT~\cite{fu2022pos} and  NPCT~\cite{guo2021pct}. Table~\ref{tab:table3} summarizes the results on ModelNet40 and ScanObjectNN datasets. Our approach outperforms all methods on ScanObjectNN dataset. 
On the ModelNet40 (synthetic dataset), our method shares the same highest accuracy with SpiderCNN, however, our method significantly outperforms SpiderCNN on  ScanObjectNN (real dataset).  
 
From the results, we can draw a few conclusions. Primarily, a pre-trained model on real 3D point cloud data provides  useful prior knowledge to help in the tasks of classification and localization for self-driving vehicles.  Secondly, most of the  methods  fail to perform well on real-world 3D point clouds even though they have higher accuracy on the synthetic datasets.  Most of the methods, except Transformer+OcCo~\cite{wang2021unsupervised} and POS-BERT ~\cite{fu2022pos},  require fully supervised learning, without leveraging pre-training, thereby   requiring large amount of labelled data. Transformer+OcCo~\cite{wang2021unsupervised} is an unsupervised Transformer approach. However, it also utilizes ModelNet40 samples for pre-training. Synthetic data for pre-training is not an optimal option for outdoor downstream tasks, e.g., localization and object detection.  Similarly, POS-BERT~\cite{fu2022pos} is a self-supervised approach, however, it is pre-trained on indoor scenes. Unlike these methods, we performed pre-training on real-world LiDAR point clouds which proved beneficial for both indoor and outdoor point cloud processing tasks.

 \begin{table*}[h]
\caption{Results on Perth-WA dataset. The values represent absolute mean error for rotation (in degrees) and additionally, the maximum error (2nd value) for translation (in meters). Our method has the least error in all cases.}
\vspace{-2mm}

\centering
\resizebox{1\textwidth}{!}{
\begin{tabular}{|l |l c  c| c|| c c c |c c |l|}

 \hline
\rotatebox[origin=c]{30}{Method}    & $Yaw$ &  $Roll$   & $Pitch$   & $Rot$  & $X$  & $Y$   & $Z$  & $Trans$  &\\ \hline
 PointLoc~\cite{wang2021pointloc} & $0.26^{\circ}$ &  $1.96^{\circ}$ & $0.153^{\circ}$ & $0.75{\circ}$, $1.51{\circ}$ & $29.70$, $83.48$ &  $37.49$, $65.98$ & $7.80$,$ 19.03$   & $25.00, 59.89$ & \\ \hline
 Ours (baseline) & $0.32^{\circ}$ & $2.42^{\circ}$ & $0.27^{\circ}$ & $1.00^{\circ}, 2.30^{\circ}$ & $ 14.20, 123.63$ & $ 17.05, 45.15$ & $8.50, 23.56$  & $13.25, 74.10$ & \\ \hline
 Ours (pretrained) &  $\mathbf{0.17^{\circ}}$ &  $\mathbf{1.52^{\circ}}$ & $\mathbf{0.096^{\circ}}$ & $\mathbf{0.59^{\circ}}$, $\mathbf{1.48^{\circ}}$ & $\mathbf{6.26, 21.44}$ & $\mathbf{ 6.55, 18.95}$ & $\mathbf{2.86, 16.64}$  & $\mathbf{5.23, 21.36} $& \\ \hline




\end{tabular}}
\label{tab:table1}

\end{table*}

\begin{table*}[h!]
\caption{Results on  Apollo-SouthBay dataset. The values represent RMSE for rotation (in rad) and translation (in meters). Our method achieves the lowest average errors.} 
\vspace{-2mm}

\centering
\resizebox{1\textwidth}{!}{
\begin{tabular}{l| c |c c c| c || c c c |l}

 \hline
\rotatebox[origin=c]{0}{Route}    & \rotatebox[origin=c]{0}{Method} & $Yaw$ &  $Roll$   & $Pitch$ & $Rot$ & $X$  & $Y$
& $Z$  & $Trans$  \\ \hline

 \multirow{3}{*}{BaylandsToSeafood} &Levinson et al.\cite{levinson2010robust} & - &  - & - & - & $0.148$ &   $0.115$ &  $0.074$ & $0.112$ \\ 
 
  & Wan et al.\cite{wan2018robust} & $\mathbf{0.054}$ & - & - & - & $0.036$ &   $\mathbf{0.026}$ &  $\mathbf{0.019}$ & $0.027$ \\ 
  
  & Ours & $0.066$ &  $0.016$ &  $0.05$ & $0.133$ & $\mathbf{0.011}$ &   $0.033$ &  $0.020$ & $\mathbf{0.021}$ \\ 
  \hline
   \multirow{3}{*}{ColumbiaPark}& Levinson et al.\cite{levinson2010robust} & - &  - & - & - & $0.063$ &   $0.045$ &  $ 0.034$ & $0.047$  \\ 
  &Wan et al.\cite{wan2018robust} & $0.081$ & - &  - & - & $0.046$ &   $0.034$ &  $0.024$ & $ 0.035$  \\ 
  & Ours& $\mathbf{0.037}$ &  $0.25$ &  $0.025$ & $0.104$ & $\mathbf{0.026}$ &   $\mathbf{0.013}$ &  $\mathbf{0.019}$ & $\mathbf{0.020}$  \\  \hline

    \multirow{3}{*}{ Hightway237} &Levinson et al.\cite{levinson2010robust} & - & - &  -& - & $0.161$ &   $0.138$ & $0.061$ & $0.120$ \\ 
    
     & Wan et al.\cite{wan2018robust}& $0.069$ & - & - & - & $0.049$ &   $\mathbf{0.038}$ &  $ 0.022$ & $0.036$ \\ 
     
      & Ours & $\mathbf{0.048}$ &  $0.23$ &  $0.118$ & $0.132$ & $\mathbf{0.013}$ &   $0.039 $ &  $\mathbf{0.014}$ & $\mathbf{0.022}$ \\ \hline
      
      \multirow{3}{*}{ MathildaAVE} &Levinson et al.\cite{levinson2010robust} &  - & - &  - & - & $0.106$ &   $0.086$ &  $0.044$ & $0.078$ \\ 
      & Wan et al.\cite{wan2018robust}&  $0.060$ & - & - &- & $0.040$ &   $0.030$ &  $\mathbf{0.020}$ & $0.030$ \\
      & Ours &  $\mathbf{0.033}$ &  $0.034 $ &  $0.190$ & $0.085$ & $\mathbf{0.019}$ &   $\mathbf{0.027}$ &  $0.039$ & $\mathbf{0.029}$ \\ \hline

        \multirow{3}{*}{SanJoseDowntown} &Levinson et al.\cite{levinson2010robust} &  - & - &  - & - & $0.103$ &   $0.075$ &  $0.055$ & $0.077$ \\ 
      & Wan et al.\cite{wan2018robust}&  $\mathbf{0.052}$ & - & - &- & $0.058$ &   $\mathbf{0.039}$ &  ${ 0.034}$ & $0.044$ \\
      & Ours &  ${0.061}$ &  $0.088  $ &  $0.147$ & $0.099$ & $\mathbf{0.054}$ &   ${0.044}$ &  $\mathbf{0.029}$ & $\mathbf{0.043}$ \\ \hline

        \multirow{3}{*}{SunnyvaleBigLoop} &Levinson et al.\cite{levinson2010robust} &  - & - &  - & - & $0.132$ &   $0.097$ &  $0.070$ & $0.099$ \\ 
      & Wan et al.\cite{wan2018robust}&  $\mathbf{0.081}$ & - & - &- & $0.069$ &   $0.050$ &  ${\mathbf{0.038}}$ & $0.052$ \\
      & Ours &  ${0.084}$ &  $0.053$ &  $0.241$ & $0.126$ & ${\mathbf{0.022}}$ &   ${\mathbf{0.043}}$ &  $0.069$ & ${\mathbf{0.045}}$ \\ \hline

       \multirow{3}{*}{\bf Average} &Levinson et al.\cite{levinson2010robust} &  - & - &  - & - & $0.119$ &   $0.093$ &  $0.046$ & $0.089$ \\ 
      & Wan et al.\cite{wan2018robust}&  ${0.066}$ & - & - &- & $0.050$ &   $0.036$ &  ${\mathbf{0.026}}$ & $0.037$ \\
      & Ours &  ${\bf 0.055}$ &  $0.117$ &  $0.128$ & $0.113$ & ${\mathbf{0.024}}$ &   ${\mathbf{0.033}}$ &  $0.031$ & ${\mathbf{0.030}}$ \\ \hline
      


\end{tabular}}
\label{tab:table2}
\vspace{-4mm}

\end{table*}

\subsection{Localization Task}
We present localization results on the proposed Perth-WA dataset and the Apollo-SouthBay dataset~\cite{L3NET_2019_CVPR}. \textcolor{black}{We did not consider KITTI~\cite{Geiger2012CVPR} dataset for experiments due to large errors in ground-truth values.} 

\subsubsection {Implementation Details} 
For fair benchmarking, we use the same implementation setup for localization in the Perth-WA dataset and the Apollo-SouthBay dataset~\cite{L3NET_2019_CVPR}. We employ batch sizes 8 and 2 for training and testing, respectively. We use Adam optimizer with learning rate 0.001 and $\beta$ values $(0.9,0.999)$. The Exponential scheduler with $\gamma = 0.9$ is used during training. The model is trained for 50 epochs on both datasets.

\subsubsection{Results on Perth-WA dataset}
Out of the 30K labeled frames, we choose 20K frames for training including sparse and dense parts of the map. For testing, an exclusive set of 10K frames is used which is not a part of the training set. \textcolor{black}{These frames were originally skipped during the construction of the training map}. The test set is challenging, in that a part of it consists of 2,200 frames that are taken from  widespread regions of Perth CBD. We evaluate the performance of a recent point cloud based localization approach PointLoc~\cite{wang2021pointloc} for comparison. We also compute baseline results that uses our method without self-supervised pretraining. 
Note that, our method is a complete localization approach even without the  self-supervised pre-trained backbone.  Following \cite{wang2021pointloc}, we use Mean Absolute Error and Max error values of poses for analysing the performance. Table~\ref{tab:table1} summarizes the results of our experiments. Our self-supervised localization approach consistently outperforms the baseline model and PointLoc for angular and translation mean error values. We can conclude from these results that our proposed self-supervised approach enables more effective  point cloud feature learning, which makes it suitable for localization using complex outdoor LiDAR frames. These results  also demonstrate that Perth-WA dataset is comparatively more challenging than the Appollo-SouthBay dataset (see Table \ref{tab:table2}) for localization.



\subsubsection{Results on the Apollo-SouthBay Dataset}
The ApolloSouthBay~\cite{L3NET_2019_CVPR} is a large scale localization dataset collected in San Francisco, USA.  An IMU based system is utilized to collect the ground-truth poses for the LiDAR frames. The dataset covers six routes, BaylandsToSeafood, ColumbiaPark, Highway237, MathildaAVE, SanJoseDowntown, and SunnyvaleBigLoop. For each route,  TrainData, TestData and MapData are provided. 
Results of our experiments on this dataset are summarized in Table~\ref{tab:table2}. We follow \cite{L3NET_2019_CVPR} and use RMSE as the evaluating metric. To establish benchmark results, we train our model on training sets, and test our model on all the routes as shown in Table~\ref{tab:table2}. We compare our approach with the state-of-the-art localization methods, Levinson et al.~\cite{levinson2010robust}  and Wan et al.~\cite{wan2018robust}. Our method outperforms both by achieving the lowest average errors. \textcolor{black}{This is mainly due to the ability of our method to exploit transformers with pre-training on large real-world data.} 

\section{CONCLUSIONS}
\vspace{-2mm}
This paper made three major contributions to outdoor localization using point cloud maps. First, it introduced a pre-text task that allows self-supervised learning, keeping in view state-of-the-art Transformer architectures. Second, it proposed a first-of-its-kind Transformer network that enables the use of multi-head attention to process outdoor LiDAR data. Third, it provided a large-scale point cloud map of Perth (Western Australia), covering nearly 4km$^2$ area. The dataset provides annotations for 6DoF localization problem and is more challenging compared to the existing Apollo-SouthBay
dataset. We established the baseline on the proposed dataset and benchmarked our method on an existing Apollo-SouthBay dataset, showing highly competitive localization results. We also established the effectiveness of our pre-text task and the model by directly fine tuning the later for classification on ModelNet40 and ScanNN datasets.

\bibliographystyle{./IEEEtran} 
\bibliography{./IEEEabrv,./IEEEexample}




\end{document}